\documentclass{article} 
\usepackage{iclr2026_conference,times}
\setcitestyle{numbers,square,comma}
\iclrfinalcopy

\usepackage{amsmath,amsfonts,bm}

\def\eqref#1{equation~\ref{#1}}

\def\1{\bm{1}}

\DeclareMathAlphabet{\mathsfit}{\encodingdefault}{\sfdefault}{m}{sl}
\SetMathAlphabet{\mathsfit}{bold}{\encodingdefault}{\sfdefault}{bx}{n}

\usepackage{hyperref}
\usepackage{url}
\usepackage[utf8]{inputenc} 
\usepackage{titlesec}       

\usepackage{graphicx}
\usepackage{xcolor}
\usepackage{subcaption}     
\usepackage[font=small,skip=5pt]{caption}
\usepackage{amsmath}  
\usepackage{amsfonts} 
\usepackage{booktabs}       
\usepackage{array}
\usepackage{tabularx}
\usepackage{makecell}       
\usepackage{multirow}       
\usepackage{adjustbox}      

\usepackage{enumitem}       
\usepackage{siunitx}        
\usepackage{lipsum}         
\usepackage[most]{tcolorbox}
\definecolor{titlebg}{RGB}{60,60,60}    
\definecolor{boxbg}{RGB}{245,245,245}
\usepackage{longtable}  
\usepackage{booktabs}   
\usepackage{tabularx}   
\usepackage{makecell}   
\usepackage{xcolor}     

\title{LegalOne: A Family of Foundation Models for Reliable Legal Reasoning}

\author{
  \parbox{0.95\linewidth}{
    Haitao Li$^{1,2}$, 
    Yifan Chen$^{1,2}$, 
    Shuo Miao$^{1,2}$, 
    Qian Dong$^{1,2}$, 
    Jia Chen$^{1,2}$, 
    Yiran Hu$^{3,\dagger}$, \\ 
    Junjie Chen$^{1,2}$, 
    Minghao Qin$^{4,\dagger}$, 
    Yueyue Wu$^{1,2}$,
    Yujia Zhou$^{1,2}$,
    Qingyao Ai$^{1,2}$\thanks{Corresponding author},  \\
    Yiqun Liu$^{1,2}$, 
    Cheng Luo$^{2,5}$, 
    Quan Zhou$^{2,5}$, 
    Ya Zhang$^{2,5}$, 
    Jikun Hu$^{2,5}$
  }
  \vspace{1em}\\
  \parbox{0.95\linewidth}{ \small
    $^1$Department of Computer Science, Tsinghua University \quad
    $^2$Quancheng Laboratory \\
    $^3$University of Waterloo, Canada \quad
    $^4$China University of Political Science and Law \\
    $^5$MegaTech.AI Inc
  }
  \vspace{0.5em}\\
  \parbox{0.95\linewidth}{\small
    \texttt{liht22@mail.tsinghua.edu.cn} 
  }
}

\begin{document}

\maketitle

\begingroup
  \renewcommand\thefootnote{$\dagger$}
  \footnotetext{Work done during internship at Tsinghua University.}
\endgroup

\thispagestyle{fancy} 
\lhead{LegalOne: A Family of Foundation Models for Reliable Legal Reasoning} 

\begin{abstract}
While Large Language Models (LLMs) have demonstrated impressive general capabilities, their direct application in the legal domain is often hindered by a lack of precise domain knowledge and complexity of performing rigorous multi-step judicial reasoning. 
To address this gap, we present LegalOne, a family of foundational models specifically tailored for the Chinese legal domain.
LegalOne is developed through a comprehensive three-phase pipeline designed to master legal reasoning. First, during mid-training phase, we propose Plasticity-Adjusted Sampling (PAS) to address the challenge of domain adaptation. 
This perplexity-based scheduler strikes a balance between the acquisition of new knowledge and the retention of original capabilities, effectively establishing a robust legal foundation.
Second, during supervised fine-tuning, we employ Legal Agentic CoT Distillation (LEAD) to distill explicit reasoning from raw legal texts. Unlike naive distillation, LEAD utilizes an agentic workflow to convert complex judicial processes into structured reasoning trajectories, thereby enforcing factual grounding and logical rigor.
Finally, we implement a Curriculum Reinforcement Learning (RL) strategy. Through a progressive reinforcement process spanning memorization, understanding, and reasoning, LegalOne evolves from simple pattern matching to autonomous and reliable legal reasoning.
Experimental results demonstrate that LegalOne achieves state-of-the-art performance across a wide range of legal tasks, surpassing general-purpose LLMs with vastly larger parameter counts through enhanced knowledge density and efficiency.
We publicly release the LegalOne weights and the LegalKit evaluation framework to advance the field of Legal AI, paving the way for deploying trustworthy and interpretable foundation models in high-stakes judicial applications.

\end{abstract}

\section{Introduction}

The demand for reliable legal AI has increased markedly in recent years~\cite{martinez2023survey,hu2026evaluation,lee2024increasing}. 
Legal professionals requires synthesizing an ever-growing set of heterogeneous sources, including statutes, regulations, administrative guidance, and case law, which exceeds what can be handled manually.
Accordingly, courts, government agencies, and law firms are beginning to adopt LLMs for document review and drafting, compliance analysis, legal research, and contract drafting, highlighting the potential of efficiency gains and lower operational costs~\cite{lai2024large,chen2024survey}.

Despite their promise, the legal domain poses unique challenges that remain tricky for general domain LLMs. 
Legal reasoning is fundamentally  \emph{knowledge-intensive}, necessitating precise grounding in authoritative sources such as statutes, judicial interpretations, and precedent. 
It is also \emph{structure-intensive}, requiring the rigorous application of multi-step legal tests and interpretive doctrines. 
In practice, these difficulties manifest in two recurring limitations: (i) insufficient coverage and understanding of structured legal knowledge, and (ii) reasoning patterns that are misaligned with real-world legal practice. As a result, when general-domain LLMs are applied to legal tasks out of the box, fluent generation does not reliably translate into legally valid, source-consistent reasoning~\cite{guha2023legalbench,li2025legalagentbench,zhang2024evaluation}.

\begin{figure*}[t]
\centering
\includegraphics[width=\linewidth]{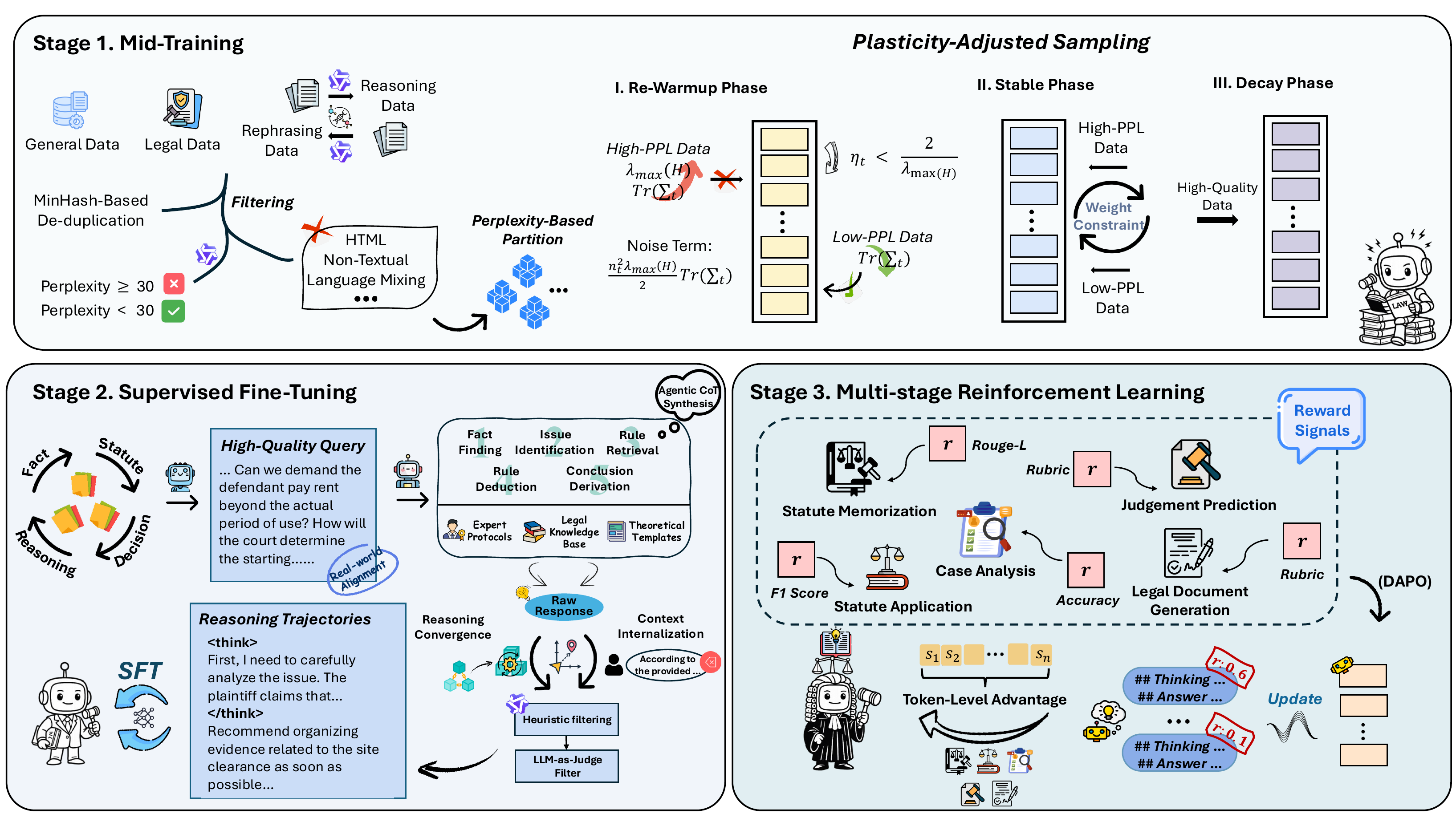}
\caption{Overview of the LegalOne Training Pipeline.}
\label{figure:overview}
\end{figure*}

To address these challenges, we present LegalOne, a series of LLMs trained specifically for the Chinese legal domain. 
LegalOne employs a multi-stage training framework designed to enhance both domain-specific knowledge and structured legal reasoning.
As shown in Figure ~\ref{figure:overview}, the pipeline comprises three distinct phases: (i) Mid-training, which constructs a robust legal semantic space through massive domain-knowledge injection; (ii) Supervised Fine-Tuning (SFT), aimed at internalizing expert priors and cultivating instruction-following capabilities; and (iii) Reinforcement Learning (RL), which catalyzes the emergence of ``Legal Mentality'' by fostering an internalized and autonomous reasoning paradigm.  
Together, these stages significantly strengthen the model’s performance on legal tasks while maintaining its general capabilities.

Specifically, during the mid-training stage, we investigate numerical stability and knowledge fidelity throughout the optimization process. 
To address these challenges, we introduce Plasticity-Adjusted Sampling (PAS), a perplexity-based data scheduling mechanism. 
This approach explicitly models the sampling probabilities of diverse data distributions as a direct function of the learning rate, enabling the data curriculum to evolve in synchronization with the model's optimization intensity.
Crucially, we maintain a fixed mixture ratio of low-perplexity ``anchor'' samples across all stages, which helps alleviate catastrophic forgetting. 
During SFT, we build an agentic system LEAD that emulates professional legal workflows, enabling the synthesis of large-scale, high-consistency reasoning trajectories. 
Finally, the RL phase implements a five-stage curriculum designed to transition the model from foundational logic to high-complexity legal scenarios. 
Through a multi-stage evolution process, LegalOne moves beyond surface-level pattern matching and demonstrates structured legal reasoning capabilities, producing autonomous, concise, and professionally grounded analyses.



\begin{figure*}[t]
\centering
\vspace{-5mm}
\includegraphics[width=0.8\linewidth]{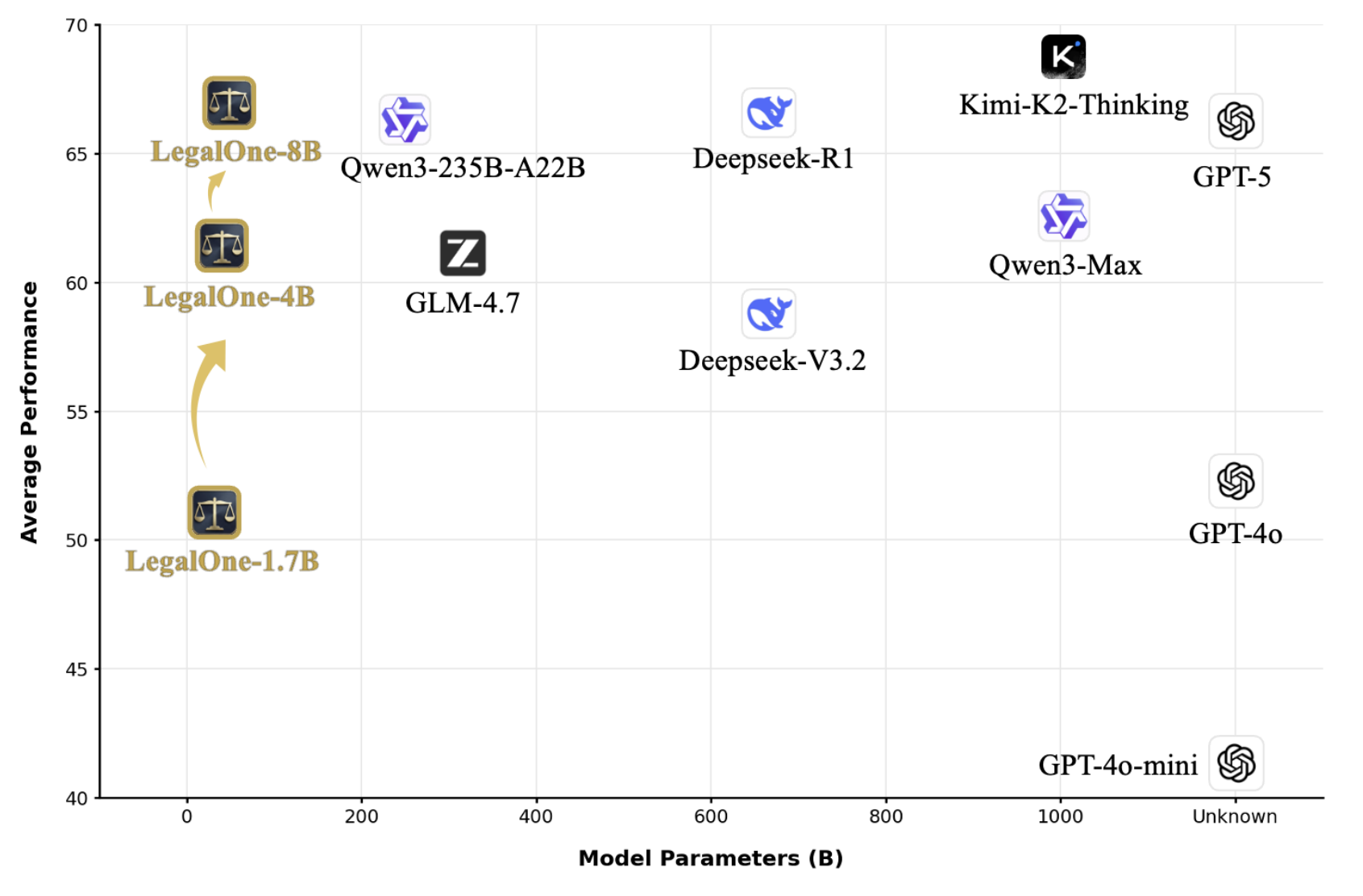}
\caption{Comparison of Model Parameters and Average Performance: The LegalOne Series Showcases Superior Parameter Efficiency.}
\label{figure:per}
\vspace{-5mm}
\end{figure*}

As shown in Figure~\ref{figure:per}, LegalOne demonstrates strong and consistent gains across a wide range of legal tasks. Crucially, it achieves superior accuracy with fewer parameters, placing it on the Pareto frontier in the performance–model-size space and dominating comparable baselines.
All models in the LegalOne series are publicly available on GitHub (\url{https://github.com/CSHaitao/LegalOne}). 
We also release LegalKit (\url{https://github.com/DavidMiao1127/LegalKit}), an evaluation toolkit designed to facilitate reproducibility and support broader research on legal-domain LLMs. 
The LegalOne series represents an important step forward in developing LLMs for the legal field. Our aim is to build trustworthy and reliable AI systems that can operate within the rigorous standards of real-world legal practice. 
We hope that the methods, insights, and resources introduced here can inform future work in legal AI and provide a blueprint for building robust, domain-aligned foundation models in other specialized fields.

\section{Mid-Training}

LegalOne introduces a dedicated mid-training stage between generic pre-training and legal instruction tuning to reduce domain shift. This stage injects high-density legal knowledge and grounds the model in legal concepts and authoritative sources, while retaining its general capabilities. This design supports accurate and reliable legal reasoning.

\subsection{Data}
The mid-training stage is supported by a carefully curated corpus that combines general data, legal data, and high-quality synthetic data.

\subsubsection{General data}

Robust legal expertise is built upon a foundation of comprehensive general knowledge. Accordingly, we curate a high-quality general-domain corpus, including datasets such as FinWeb-Edu~\cite{lozhkov2024fineweb-edu}, FinePDFs~\cite{kydlicek2025finepdfs}, FineWiki~\cite{penedo2025finewiki}, SkyPile-150B~\cite{wei2023skywork}, IndustryCorpus~\cite{beijing_academy_of_artificial_intelligence}, OpenNewsArchive~\cite{he2024opendatalabempoweringgeneralartificial}, MathPile~\cite{wang2024mathpilebilliontokenscalepretrainingcorpus}, Wanjuan~\cite{he2023wanjuancomprehensivemultimodaldataset}, and BaiduBaike-5.63M.
This general data serves as a critical regularization mechanism, alleviating catastrophic forgetting as the model undergoes legal specialization.

As perfectly preserving all capabilities is inherently difficult, we adopt a task-centric strategy for mid-training.  
We prioritize advancing legal reasoning, while deprioritizing less relevant capabilities like advanced coding.
This selective focus ensures that LegalOne’s learning capacity is effectively allocated toward achieving the reliability and precision required by the legal field.

\subsubsection{Legal Data}

We constructed a large-scale legal corpus by aggregating heterogeneous data from public repositories and professional legal sources. 
The corpus covers a broad range of legal discourse, including legal academic papers, law review articles, commentaries, consultations, court judgments, judicial opinions, textbooks, encyclopedias, guiding opinions, and statutory and regulatory texts. 

For PDF-only materials (e.g., academic publications and legal textbooks), we adopt a rigorous data processing approach that begins with OCR-assisted extraction. The extracted content is then parsed and normalized using regular expressions, followed by automated cleaning and targeted manual verification of key fields to ensure the resulting data is high-quality and machine-readable.

We prioritize data with high educational value for legal comprehension and reasoning. 
To ensure temporal relevance and mitigate noise from deprecated norms, we selectively retain legal documents from the past five years and prioritize statutory provisions with current normative validity. 
Furthermore, we implement an automated quality gating process for web-sourced content; specifically, we employ the fineweb-edu-scorer~\cite{lozhkov2024fineweb-edu} to evaluate data quality, systematically filtering out samples with a score below 3 to maintain a high-fidelity training corpus.

\subsubsection{Synthetic Data}
Synthetic legal data forms an important component of our mid-training corpus. We focus on two main categories of synthesized data: reasoning data and rephrasing data.

\textbf{Reasoning Data.}
Prior work shows that incorporating explicit reasoning signals during pretraining can significantly enhance downstream performance~\cite{akter2025front,jiang2025rationalyst}. 
However, a critical gap exists in real-world corpora: while judicial and scholarly documents are intellectually dense, their underlying logic is often implicit or fragmented. Consequently, such documents lack transparent logical chains and multi-stage syllogisms, which are crucial for cultivating rigorous legal reasoning capabilities.
To address this limitation, we generate synthetic reasoning samples from high-quality legal sources such as legal academic papers, authoritative commentaries, and legal textbooks. 
We leverage Qwen3-255B-A22B-Thinking~\cite{yang2025qwen3} to generate structured reasoning trajectories that mirror professional legal discourse. 
Inspired by the Feynman Technique~\cite{reyes2021feynman}, we guide the model to distill complex doctrinal principles into accessible, granular analytical steps.

\textbf{Rephrasing Data.}
Rephrasing has been shown to improve pre-training efficiency by increasing the knowledge learned per token~\cite{team2025kimi}. 
To enhance linguistic diversity while preserving factual accuracy, we adopt an approach similar to Kimi-K2~\cite{team2025kimi}. 
Using carefully designed prompts, we instruct Qwen3-255B-A22B-Thinking~\cite{yang2025qwen3} to generate faithful paraphrases of source texts across different writing styles and perspectives. 
We then assess semantic consistency between each paraphrase and its original text using a vector-based similarity model, and remove samples that exhibit low semantic alignment.

\subsubsection{Data Filtering}
To ensure the integrity of the mid-training corpus, we implement a multi-stage data refinement and filtering pipeline designed to maximize information density while minimizing semantic noise.

\textbf{De-duplication.} We apply MinHash similarity detection to remove near-duplicate documents across all sources, preventing overfitting to repeated patterns and improving training efficiency.

\textbf{Heuristic filtering.} 
For datasets with pre-existing quality metrics, we primarily sample from higher-scoring subsets.   
Our cleaning pipeline further includes two key steps: structural normalization (which strips HTML tags and non-textual elements) and the removal of documents below a minimum length threshold (e.g., 200 characters).
For synthetic corpora, we remove samples with excessive mixed-language or stylistic drift that may compromise training stability.

\textbf{Perplexity-based filtering.}
To establish a unified quality metric across heterogeneous data sources, we compute the perplexity (PPL) of each sample using the Qwen-3-4B-base model. 
Our empirical analysis reveals a clear correlation between high perplexity and diminished data quality. 
Specifically, documents with PPL values exceeding 30 are frequently characterized by linguistic noise or poor distributional alignment with our training objectives. 
Consequently, we adopt 30 as the threshold and filter out samples that exceed this limit to ensure a high-fidelity training corpus.

\textbf{Language constraints.}
Since LegalOne is designed specifically for the Chinese legal domain, we retain only Chinese legal documents for domain-specific training. 
For general-domain corpora, we keep a curated subset of high-quality English texts to preserve basic multilingual reasoning capabilities.


Through this multi-step filtering and selection pipeline, we obtain approximately 100B tokens of high-quality data for mid-training.
During mid-training, we maintain a strategic ratio of 4:6 between general and legal-domain data, ensuring a balance between broad linguistic intelligence and deep domain expertise. 
This high-fidelity dataset serves as a clean, heterogeneous foundation, providing a stable representational base for LegalOne’s subsequent training phases.

\subsection{Perplexity-based Data Scheduling}

\subsubsection{Motivation}

During the mid-training phase, our primary objective is to facilitate the acquisition of specialized legal knowledge while mitigating catastrophic forgetting of pre-existing capabilities. 
Prior work has shown that learning rate scheduling and data replay play a crucial role in balancing new and old knowledge~\cite{mo2025mid,ibrahim2024simple}.
However, in practice, we are limited to resume training from an existing open-source checkpoint, which introduces two additional challenges.

First, the absence of original pre-training data complicates the construction of an effective replay buffer. 
To bridge this gap, we adopt a heuristic operational assumption: samples exhibiting lower perplexity (PPL) under the current base model are more likely to align with its original latent distribution. 
Consequently, these low-PPL samples serve as a functional proxy for the unknown pre-training data, providing a stabilizing mechanism that approximates the effects of data replay.
Second, open-source base models are typically released at the terminal stage of a decay schedule, where the learning rate (LR) has reached a low, stable regime. 
If we continue mid-training directly at this low learning rate, the model struggles to effectively absorb the large amount of newly introduced legal-domain knowledge.
To restore model plasticity and improve learning efficiency, it is necessary to re-warm the LR to a higher peak. 
However, such re-warming can induce optimization shock, leading to catastrophic forgetting~\cite{gupta2023continual,wang2025learning}. 
Thus, a key challenge in mid-training is how to smoothly raise the learning rate to the target range without significantly harming the model’s existing knowledge.

We argue that learning-rate scheduling and data distribution control should be treated as a coupled problem, rather than two independent techniques. 
Intuitively, large shifts in learning rate and data distribution occurring simultaneously can destabilize optimization. 
We therefore introduce data scheduling to explicitly control how the training distribution evolves over time, coordinating it with the learning-rate schedule. 
Concretely, we use the perplexity (PPL) as a lightweight measure of distribution proximity: lower PPL generally indicates closer alignment with the model’s current data manifold. 
Based on this signal, we propose perplexity-based data scheduling to adjust the training mix in a stable, model-aware manner. 
We detail the method below.

\subsubsection{Theoretical Analysis}

To understand the impact of data scheduling on model evolution during mid-training, we analyze the optimization process from the perspectives of \textit{numerical stability} and \textit{knowledge fidelity}.

\textbf{Numerical Stability.}
To formally analyze the stability during the re-warmup phase, we consider the $L$-smoothness of the loss function $\mathcal{L}(\theta)$. The change in loss after a parameter update $\theta_{t+1} = \theta_t - \eta_t \tilde{g}_t$, where $\eta_t$ denotes the learning rate at step $t$, can be characterized by the second-order Taylor expansion:
\begin{equation}
\mathcal{L}(\theta_{t+1}) - \mathcal{L}(\theta_t) \approx -\eta_t \nabla \mathcal{L}(\theta_t)^\top \tilde{g}_t + \frac{\eta_t^2}{2} \tilde{g}_t^\top \mathbf{H}(\theta_t) \tilde{g}_t
\end{equation}
where $\mathbf{H}(\theta_t)$ is the Hessian matrix $\nabla^2 \mathcal{L}(\theta_t)$. Taking the expectation over the stochastic gradient $\tilde{g}_t$ (where $\mathbb{E}[\tilde{g}_t] = \nabla \mathcal{L}(\theta_t)$) and utilizing the property $\mathbf{v}^\top \mathbf{H} \mathbf{v} \leq \lambda_{max}(\mathbf{H}) \|\mathbf{v}\|^2$, we obtain:
\begin{equation}
\mathbb{E}[\mathcal{L}(\theta_{t+1}) - \mathcal{L}(\theta_t)] \leq -\eta_t \|\nabla \mathcal{L}(\theta_t)\|^2 + \frac{\eta_t^2 \lambda_{max}(\mathbf{H})}{2} \mathbb{E}[\|\tilde{g}_t\|^2]
\end{equation}

By decomposing the second moment of the gradient into its mean and variance, $\mathbb{E}[\|\tilde{g}_t\|^2] = \|\nabla \mathcal{L}(\theta_t)\|^2 + \text{Tr}(\Sigma_t)$, where $\text{Tr}(\Sigma_t)$ represents the gradient covariance (noise variance), the inequality can be rearranged as:
\begin{equation}
\mathbb{E}[\mathcal{L}(\theta_{t+1}) - \mathcal{L}(\theta_t)] \leq -\eta_t \left( 1 - \frac{\eta_t \lambda_{max}(\mathbf{H})}{2} \right) \|\nabla \mathcal{L}(\theta_t)\|^2 + \frac{\eta_t^2 \lambda_{max}(\mathbf{H})}{2} \text{Tr}(\Sigma_t)
\end{equation}

Equation (3) reveals that expected loss reduction requires a delicate balance between the learning rate $\eta_t$ and the Hessian spectrum. Specifically, the stability condition $\eta_t < 2/\lambda_{max}(\mathbf{H})$ must be maintained to prevent second-order curvature effects from overwhelming the gradient descent term.

During the re-warmup phase, the quadratic noise term $\frac{\eta_t^2 \lambda_{max}(\mathbf{H})}{2} \text{Tr}(\Sigma_t)$ becomes the primary source of instability. 
For high-PPL data, its out-of-distribution (OOD) nature triggers a simultaneous surge in both local curvature $\lambda_{max}(\mathbf{H})$ and gradient variance $\text{Tr}(\Sigma_t)$, significantly narrowing the stability window. 
In contrast, low-PPL data aligns with the model's current manifold, minimizing $\text{Tr}(\Sigma_t)$ and maintaining a benign curvature profile. 
Consequently, low-PPL data acts as numerical damper, stabilizing the training process against the turbulence caused by increasing $\eta_t$.

Thus, we minimize distribution shifts during learning rate expansion to preserve numerical stability. 
The inclusion of low-PPL data anchors the parameter trajectory within a stable manifold until the model achieves sufficient momentum for the subsequent stable phase.

\textbf{Knowledge Fidelity.}
After the re-warmup phase, the optimizer's first and second moments have effectively encoded the statistical manifold of the original distribution. 
This momentum memory establishes a robust equilibrium, enabling the model to absorb high-PPL domain knowledge without collapsing.

Therefore, in this Stable phase, our primary objective shifts to balancing the acquisition of new knowledge with the retention of historical data. 
To quantify this, we analyze the evolution of the validation loss $\mathcal{L}_{val}$ on $\mathcal{D}_{pre}$. 
Following an update $\Delta \theta = -\eta_{max} \mathbf{g}_{train}$, the variation of validation loss $\Delta \mathcal{L}_{val}$ is approximated by:
\begin{equation} 
\Delta \mathcal{L}_{val} \approx -\eta_{max} \underbrace{\langle \nabla \mathcal{L}_{val}, \mathbf{g}_{train} \rangle}_{\text{Alignment Term}} + \frac{\eta_{max}^2}{2} \underbrace{\mathbf{g}_{train}^\top \mathbf{H}_{val} \mathbf{g}_{train}}_{\text{Curvature Term}} 
\end{equation}
where $\mathbf{H}_{val}$ is the Hessian matrix of the validation set. 
To minimize catastrophic forgetting (i.e., keeping $\Delta \mathcal{L}_{val}$ as small as possible), the training gradient $\mathbf{g}_{train}$ must maintain a positive projection onto $\nabla \mathcal{L}_{val}$ while suppressing excessive curvature energy. 
At this stage, relying solely on high-PPL data is risky; such gradients typically exhibit severe directional drift (a negative inner product with $\nabla \mathcal{L}_{val}$) and high curvature energy, may leading to a sharp rise on $\Delta \mathcal{L}_{val}$.

Consequently, maintaining a strategic proportion of low-PPL data during the Stable phase is essential to anchor the optimization. 
The gradients from this data serve a dual purpose: they steer new learning in a way that protects what the model already knows, while also helping to integrate new knowledge without conflict.

\subsubsection{Preliminary Experiment}
To empirically validate the theoretical analysis, we conducted a preliminary mid-training study using the Qwen3-0.6B-Base model on a legal corpus.

Specifically, we stratified a 10B-token legal corpus, $\mathcal{D}_{legal}$, into three disjoint subsets based on their perplexity (PPL) relative to the base model: $\mathcal{D}_{low}$ (3B tokens, $PPL \leq 5$), $\mathcal{D}_{mid}$ (4B tokens, $5 < PPL \leq 15$), and $\mathcal{D}_{high}$ (3B tokens, $PPL > 15$). This partitioning effectively quantifies the degree of domain shift, with higher PPL indicating samples that reside further from the model's original knowledge manifold.

Following \cite{ibrahim2024simple}, we monitor catastrophic forgetting by evaluating the cross-entropy loss on a held-out validation set, $\mathcal{D}_{val}$, every 100 training steps. $\mathcal{D}_{val}$ is sampled from Chinese Wikipedia, which is widely used in large-scale pretraining corpora and thus serves as a stable proxy for the model’s original distribution. 
All experiments utilize a fixed budget of 10B tokens and identical hyperparameters.
We employ a WSD (Warmup-Stable-Decay)~\cite{hu2024minicpmunveilingpotentialsmall} learning rate scheduler with a peak learning rate of 3e-4. Specifically, the first 10\% of total steps are dedicated to a re-warmup phase to ensure stable optimization. Based on this framework, we compare four training schedules:

\textbf{Static Mixture.} Throughout the entire training process, each step samples data from a fixed distribution, maintaining a constant ratio of $\mathcal{D}_{low} : \mathcal{D}_{mid} : \mathcal{D}_{high} = 3 : 4 : 3$.

\textbf{Ascending Curriculum.} The model encounters the subsets sequentially according to their distributional complexity: $\mathcal{D}_{low} \rightarrow \mathcal{D}_{mid} \rightarrow \mathcal{D}_{high}$.

\textbf{Descending Curriculum.} A direct inversion of the ascending schedule, where the model is first exposed to the most complex samples: $\mathcal{D}_{high} \rightarrow \mathcal{D}_{mid} \rightarrow \mathcal{D}_{low}$.

\textbf{Two-Stage Hybrid.} This schedule bifurcates the distribution based on the re-warmup phase. During the first 10\% of steps, the model is trained exclusively on $\mathcal{D}_{low}$ to stabilize the initial optimization. For the remaining 90\%, it transitions to a fixed mixture with a ratio of $\mathcal{D}_{low} : \mathcal{D}_{mid} : \mathcal{D}_{high} = 2:4:3$.

\begin{figure*}[t]
\centering
\includegraphics[width=0.8\linewidth]{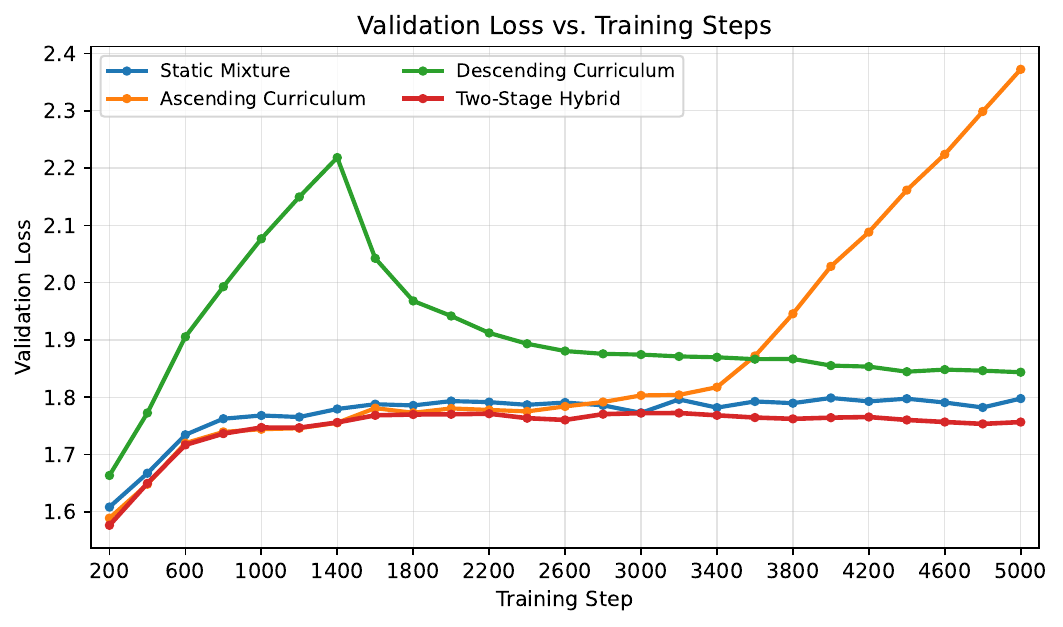}
\caption{Comparison of validation loss across different training schedules during 10B-token continual pre-training.}
\label{figure:loss}
\vspace{-5mm}
\end{figure*}

Figure~\ref{figure:loss} illustrates the results of these preliminary experiments, revealing varying degrees of catastrophic forgetting induced by different data curricula. Our key observations are summarized below.

\textbf{Decoupling of Training and Validation Dynamics.} Training and validation losses often exhibit divergent trajectories. While training loss decreases consistently, validation loss on historical distributions ($\mathcal{D}_{val}$) may surge. This decoupling suggests that training loss alone is an inadequate proxy for knowledge retention. Thus, monitoring validation loss is essential to accurately detect catastrophic forgetting.

\textbf{Risk of Isolated High-Complexity Data.}
Exposing the model exclusively to high-PPL (complex) data triggers irreversible forgetting. Training on such out-of-distribution samples in isolation forces parameters to drift away from the original optimization basin. To mitigate this, we recommend integrating a fixed proportion of low-PPL data as an ``anchor'' in every stage to maintain spatial stability and prevent drastic parameter shifts.

\textbf{Sensitivity of the Re-warmup Phase.} The re-warmup phase is uniquely vulnerable to instability, as evidenced by significant fluctuations in validation loss. To counter this, we recommend using data that closely aligns with the model’s original distribution during this stage. Specifically, prioritizing low-PPL data acts as a ``buffer'', stabilizing the loss landscape as the learning rate scales and protecting the model from early-stage divergence.

\subsubsection{Method}
Based on our theoretical analysis, we propose Plasticity-Adjusted Sampling (PAS). This method treats the sampling probability as a direct function of the learning rate, allowing the data distribution to evolve in sync with the model's optimization intensity.
We partition the mid-training corpora into $N$ buckets based on perplexity (PPL). Let $\alpha(t) = \eta(t) \eta_{max} \in [0, 1]$ be the Plasticity Coefficient at step $t$. The sampling probability $P(\mathcal{B}_i | t)$ for each bucket $\mathcal{B}_i$ is defined as:

\begin{equation}
P(\mathcal{B}_i | t) = \frac{w_i \cdot \exp\left( \lambda \cdot \mathbb{I}_{i=1} \cdot (1-\alpha(t)) \right)}{\sum_{j=1}^N w_j \cdot \exp\left( \lambda \cdot \mathbb{I}_{j=1} \cdot (1-\alpha(t)) \right)}
\end{equation}

where $w_i$ is the target equilibrium distribution for stable training, $\mathbb{I}_{i=1}$ is an indicator function anchoring the lowest-PPL bucket, and $\lambda$ (set to 5) governs the initial concentration intensity.

The PAS mechanism ensures a smooth transition through two phases. At the start of re-warmup ($\alpha \to 0$), the exponential term $\exp(\lambda)$ acts as a powerful prior. 
This forces the model to focus almost exclusively on familiar low-PPL anchor data, stabilizing representations while the learning rate rises. 
As the learning rate peaks ($\alpha \to 1$), the bias term $(1-\alpha(t))$ vanishes, and the exponential multiplier reverts to unity ($e^0 = 1$). 
The distribution then seamlessly relaxes into the predefined target mixture $w$, where the low-PPL data settles at its steady-state proportion of 20\%.

\subsubsection{Training details}
For the mid-training stage of LegalOne, we utilize an Ascend 910B cluster powered by the MindSpeed-LLM framework, adopting the Megatron-Core parallelism paradigm to facilitate large-scale distributed training. To maintain architectural consistency with the final training phase of Qwen-3-Base, the context window is fixed at 32K tokens. We implement sample packing to maximize throughput and data efficiency, resulting in an effective global batch size of approximately 4M tokens.

Optimization is performed using the AdamW optimizer with hyperparameters $\beta_1 = 0.9$ and $\beta_2 = 0.95$, complemented by a weight decay of $0.1$ and a gradient clipping threshold of $1.0$ to safeguard numerical stability. The learning rate follows a Warmup-Stable-Decay (WSD) schedule: it is linearly increased over 2,000 steps to a peak value of $3 \times 10^{-4}$, maintained at this plateau during the stable training phase, and finally annealed to $3 \times 10^{-6}$ during the cooldown phase using a targeted 10B token dataset. Following Hu et al.~\cite{hu2024minicpmunveilingpotentialsmall}, we incorporate the entirety of our SFT data into this annealing phase to further align the model with high-quality instruction-following distributions.

To handle long-context dependencies, we employ Rotary Positional Embeddings (RoPE) ~\cite{su2024roformer} with an expanded base of 1,000,000. Furthermore, training efficiency and numerical robustness are enhanced through the integration of FlashAttention~\cite{dao2022flashattention} and RMSNorm~\cite{zhang2019root}, ensuring high-performance attention computation and consistent layer normalization across the distributed architecture.

\section{Legal Agentic CoT Distillation (LEAD)}

During the supervised fine-tuning(SFT) stage, 
it is standard to rely on a stronger general-purpose teacher LLM to produce CoT traces, which are subsequently used to supervise the student model. 
However, in legal domain, even the most advanced general-purpose models exhibit notable limitations in factual accuracy, legal applicability, and the rigor of their arguments. Naively distilling their outputs may propagate and amplify these errors and biases, posing a serious threat to the reliability and safety of the legal LLM.

Meanwhile, real-world legal documents serve as an invaluable repository of authoritative social facts and structured legal logic. 
However, these records are primarily post-hoc summaries tailored for professional audiences, such as judges and attorneys. 
As for post-hoc summaries for professionals, they frequently omit critical intermediate steps, such as identifying disputed issues or applying legal provisions. 
This lack of explicit trajectories prevents LLMs from recovering the underlying logic, thereby hindering the acquisition of transferable reasoning skills.

\begin{figure*}[t]
\centering
\includegraphics[width=\linewidth]{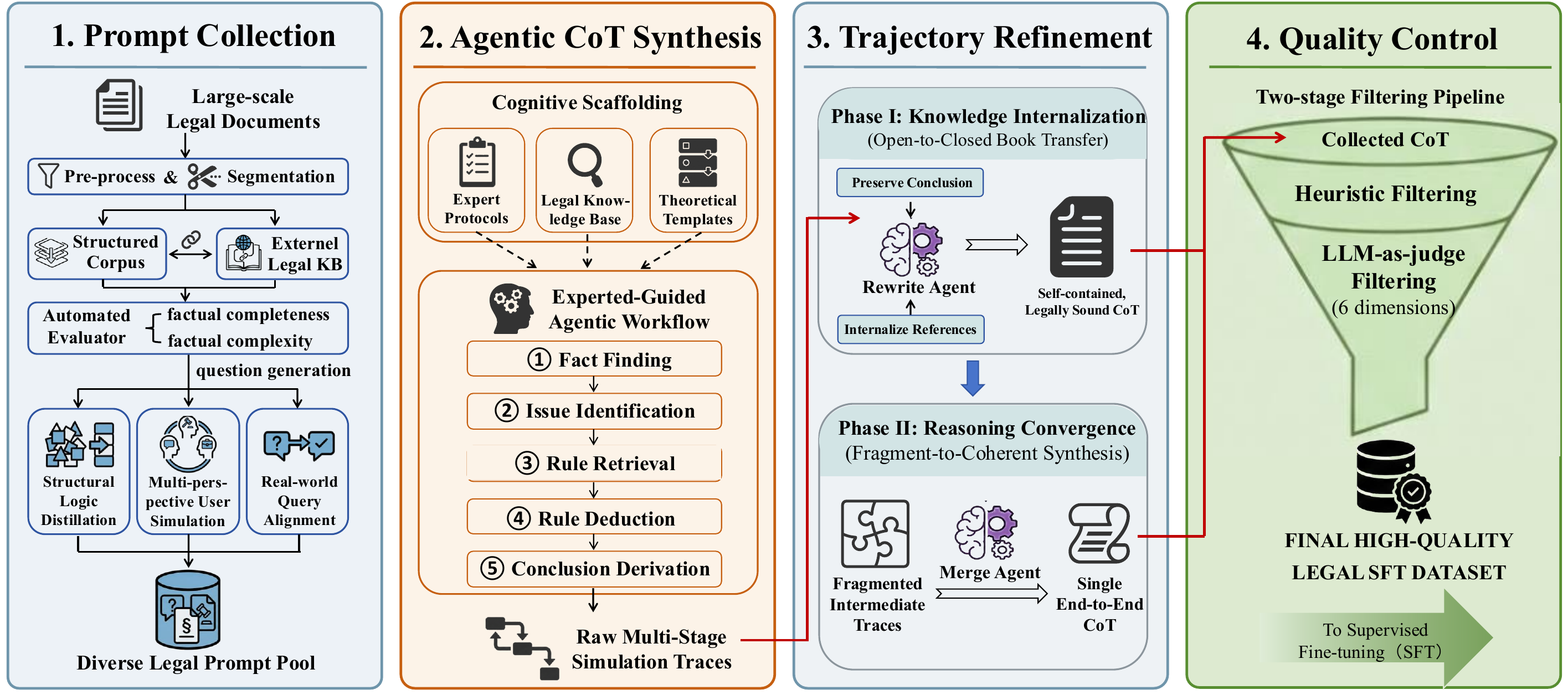}
\caption{The architecture of the proposed Legal Agentic CoT Distillation (LEAD) pipeline.}
\label{figure:sft}
\vspace{-5mm}
\end{figure*}

To bridge this gap, we propose LEAD (Legal Agentic CoT Distillation). 
As illustrated in Figure ~\ref{figure:sft}, the pipeline comprises four distinct stages: Prompt Collection, Agentic COT Synthesis, Trajectory  Refinement and Quality Control. 
The core philosophy of our system lies in deconstructing complex legal reasoning into structured, real-world workflows. 
By leveraging agentic orchestration, we enable general-purpose LLMs to execute tasks in a modular fashion, thereby generating reasoning trajectories that are factually grounded, logically rigorous, and closely aligned with judicial practice. 
The following sections provide a detailed exposition of this process.

\subsection{Prompt Collection}
\subsubsection{Structured Case Corpus Construction}

We construct a high-quality base corpus by curating a large-scale collection of judicial documents issued within the last five years.
To ensure data integrity, we first implement a pre-processing pipeline to exclude outliers, such as documents with excessive length (e.g., those containing voluminous annexes) or those lacking substantive content (e.g., brief procedural rulings).

For the retained documents, we employ heuristic regex-based segmentation tailored to the standardized format of Chinese judicial writing. 
This allows us to decompose each document into three functional modules: \textit{Fact}, \textit{Reasoning}, and \textit{Decision}. 
Cases that fail this structural parsing are pruned to maintain high-quality supervision signals. 
Furthermore, we extract cited statutes and link them to an external legal knowledge base, creating a mapping between factual patterns and applicable legal provisions. 
This structured knowledge serves as the grounding context for our subsequent pipeline.

To guarantee the instructional quality of the corpus, we employ Qwen3-235B-A22B-Thinking~\cite{yang2025qwen3} as an automated evaluator. Each case is assessed on a 1--5 scale across two dimensions: \textbf{factual completeness} and \textbf{factual complexity}. 
We exclude cases with scores below 3 to ensure the model learns from sufficiently informative examples.
Finally, to counteract the imbalanced distribution of causes of action, we apply a stratified sampling strategy. 
By downsampling prevalent case types and oversampling sparse categories, we achieve a balanced and diverse corpus that prevents the model from developing frequency-induced biases.

\subsubsection{Question Generation} 

To transform the static corpus into an interactive training set, we design a multi-faceted task generation framework:

\textbf{Structural Logic Distillation.} We exploit the inherent logical dependencies between the four components: \textit{Fact}, \textit{Reasoning}, \textit{Decision}, and \textit{Legal Provisions}. By treating different sections as inputs or targets, we prompt the LLM to: (i) derive the reasoning process from facts, (ii) analyze the applicability of provisions to specific fact patterns, and (iii) predict the final judgment. These tasks align with the cognitive workflow of legal practitioners, forcing the model to internalize step-by-step judicial logic.

\textbf{Multi-perspective User Simulation.} To capture the diverse needs of different stakeholders, we employ an LLM to simulate queries from roles such as litigants, attorneys, and judges. Using the \textit{Fact} section as context, the model generates realistic tasks, including legal advice requests, risk assessments, and document drafting. Since the underlying corpus covers a wide range of case types and procedural stages, these simulated questions naturally exhibit high diversity in both intent and form. This mechanism expands the task space beyond document-centric reasoning, enabling the model to handle open-ended, user-facing queries while remaining grounded in factual evidence.

\textbf{Real-world Query Alignment.} To bridge the gap between synthetic prompts and actual user behavior, we collect real-world queries from online legal consultation platforms. These inputs capture nuances of authentic human interaction, including vague descriptions, ambiguous references, and fragmented clarifications that are often absent from model-generated data.

\subsection{Agentic CoT Synthesis.}

Existing LLMs exhibit a critical limitation, which we refer to as procedural agnosia. Despite possessing vast parametric knowledge, they inherently lack the \textit{practical methodologies} derived from real-world judicial practice. Specifically, they fail to grasp the tacit Standard Operating Procedures followed by legal professionals. To address this, we propose the \textbf{Agentic COT Synthesis} framework designed to simulate the authentic cognitive process of legal experts. By collaborating with senior practitioners, we formalize abstract judicial reasoning into a structured, explicit agentic workflow.

We define the reasoning architecture as a configurable workflow topology that breaks down complex legal problems into a network of autonomous Agentic Processes. Each cognitive state (e.g., fact-finding, rule deduction) functions as an autonomous decision node capable of proactively querying external databases for targeted knowledge retrieval~\cite{li2023sailer,shi2025deep}.

As shown in Figure \ref{figure:sft}, we instantiate specialized judicial state machines for different legal tasks. Taking judgment prediction as an example, the workflow is decomposed into five stages: Fact Finding, Issue Identification, Rule Retrieval, Rule Deduction, and Conclusion Derivation. By transforming legal documents into transparent, expert-aligned execution graphs, the framework enhances solution accuracy while enabling the model to acquire explicit reasoning paths.

To uphold legal professional standards and integrate realistic problem-solving logic, every node is supported by a comprehensive Cognitive Scaffold.

\begin{itemize} 

\item \textbf{Expert Protocols.} For specific tasks, expert protocols defined by legal professionals, including standard procedures and common pitfalls, are injected directly into the LLM’s context. These protocols act as explicit guardrails, guiding the agent to strictly adhere to actual courtroom standard operating procedures, thereby preventing common procedural errors.

\item \textbf{Legal Knowledge Base.} To enforce evidence-based reasoning, each node integrates an external retrieval module that can recalls statutes and precedents to substantiate judgments. For critical decision nodes, we implement pre-retrieval and re-ranking strategies to ensure the injection of highly relevant legal context.

\item \textbf{Theoretical Templates.} We utilize doctrinal frameworks (e.g., judicial syllogism or four-element analysis) to structure the agent's reasoning. This constraint transforms the Chain-of-Thought (CoT) from free-form generation into a rigorous ``major premise–minor premise–conclusion'' argument, effectively preventing logical leaps and hallucinations.

\end{itemize}

\subsection{Trajectory Refinement}

While the agentic workflow generates detailed intermediate reasoning, the raw traces differ significantly from ideal training targets. To address this, we introduce a two-phase refinement process, Knowledge Internalization and Reasoning Convergence, to distill these raw traces into robust training data.

\subsubsection{Knowledge Internalization}

A key challenge is the information gap between the teacher and student models. During the Agentic CoT synthesis phase, the model operates in an ``open-book'' mode, having access to retrieved statutes, expert protocols, and theoretical templates within the context.

However, during student model training, we do not want the model to rely on this rich context. We want it to answer questions directly, as real-world users rarely provide such detailed ``cognitive scaffolding.'' This shift in data distribution can lead to hallucinated citations or over-reliance on context that is not available during deployment.

To mitigate this issue, we introduce a knowledge internalization stage. Given the initially generated chain-of-thought and final answer, we prompt a strong LLM to rewrite the content with minimal changes. The constraints are: (i) the final conclusion and high-level reasoning structure must be preserved, but (ii) all explicit dependence on external references must be paraphrased as if it arises from the model’s internal knowledge. Specifically, the model is required to remove direct citations and phrases that reveal external sources (e.g., ``according to the reference materials''), rewriting them as its own analysis. This task is straightforward for current strong LLMs. This step produces self-contained, legally sound chains of thought aligned with real-world inputs, enabling the student model to learn robust reasoning patterns without hidden context, thereby reducing hallucinations.

\subsubsection{Reasoning Convergence}
While the multi-stage workflows above substantially reduce the difficulty of complex legal tasks, they also yield local chains of thought, each tied to a specific substep. Because these intermediate traces are produced under different prompts and contexts, they are not guaranteed to form a globally coherent argument and may exhibit repetition or unfinished sub-analyses. Therefore, simply concatenating them into a single training chain of thought is ill-advised.
To address this, we introduce a \emph{reasoning convergence} step that that merges the stepwise traces into a single end-to-end chain of thought.
Given the sequence of intermediate rationales and the final conclusion, we prompt a strong LLM to merge them into a single chain that (i) preserves the correctness of the final outcome and the key reasoning steps from each stage, (ii) maintains global coherence and logical order, and (iii) removes redundant or tangential analysis.
The result of this process is a compact, self-consistent reasoning trajectory that faithfully reflects the underlying multi-step workflow, but is presented as a single, human-readable legal argument.

\subsection{Quality Control}

After knowledge internalization and reasoning convergence, we perform a two-stage filtering process to further improve the reliability and usability of the collected data.

\paragraph{Heuristic filtering.}
We first apply lightweight heuristic filters to remove clearly defective samples. 
We discard responses that are truncated or structurally incomplete, remove samples with excessive Chinese–English code-mixing beyond a preset threshold, and eliminate clearly duplicated training instances. 
This inexpensive step removes obvious noise and ensures that only well-formed examples proceed to the more costly quality assessment stage.

\paragraph{LLM-as-judge filtering.}
In the second stage, we use an LLM-as-judge~\cite{li2024llms,li2025calibraeval} to perform fine-grained evaluation of the remaining samples. For each (prompt, reasoning, answer) triple, a general LLM assigns 1–10 scores on several quality dimensions:
\begin{enumerate}
    \item \textbf{Reasoning quality}: logical validity and completeness of the chain-of-thought, including coverage of key inferential steps and legal soundness.
    \item \textbf{Reasoning consistency}: maintaining a stable argumentative position throughout the reasoning process, without internal contradictions or unexplained shifts in perspective, assumptions, or conclusions.
    \item \textbf{Answer-reasoning consistency}: alignment between the final answer and the preceding reasoning, such that the conclusion neither contradicts nor selectively ignores the chain of thought.
    \item \textbf{Conciseness}: absence of excessive repetition, digressions, or irrelevant analysis, while retaining essential legal arguments.
    \item \textbf{Linguistic}: fluency and stylistic coherence, including stable use of language and consistent legal style.
    \item \textbf{Overall score}: an overall judgment of how well the sample serves as pedagogical supervision for legal reasoning.
\end{enumerate}
We then discard all samples whose overall score or any subscore falls below 7, thereby removing cases with weak reasoning, poor consistency, or limited training value. 
The remaining high-scoring samples constitute our final legal-domain SFT dataset, which is structurally clean and rigorously vetted for reasoning quality. 
To ensure general capabilities, we also incorporate high-quality open-source datasets, resulting in a final SFT dataset of 500k samples. 
We set the maximum learning rate to 2e-5 and train the model for 2 epochs, with a batch size of 32 and a context length of 32k.

\begin{figure*}[t]
\centering
\includegraphics[width=\linewidth]{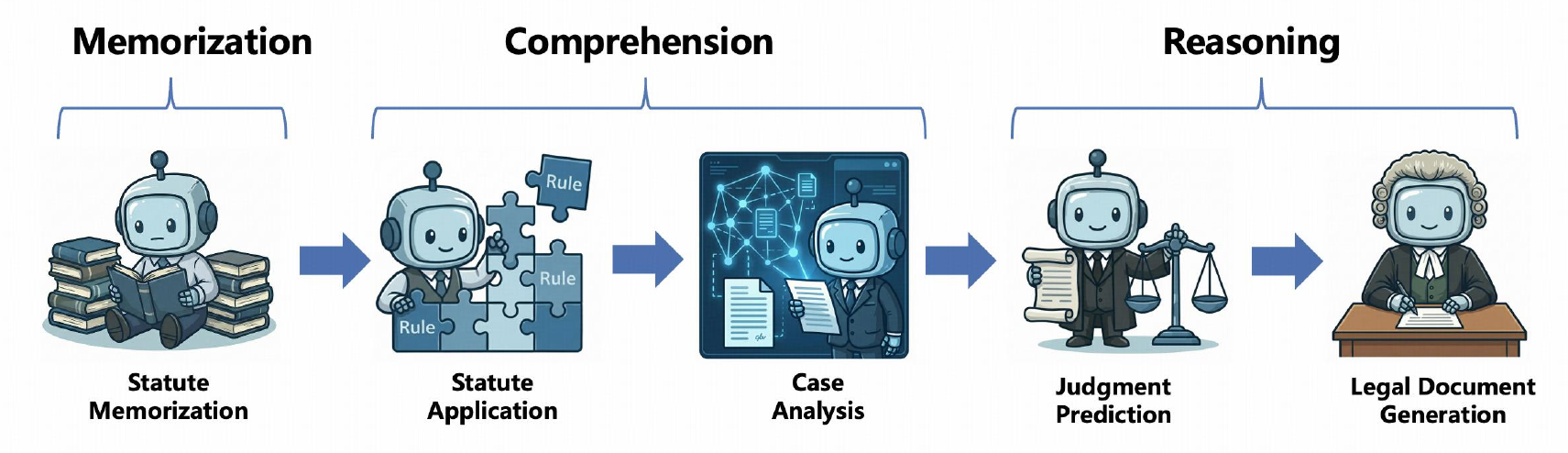}
\caption{Overview of the Multi-Stage Reinforcement Learning Framework for LegalOne.}
\label{figure:RL}
\end{figure*}

\section{Multi-stage Reinforcement Learning}

After mid-training and SFT, the LLM has acquired a broad base of legal knowledge and an initial capability for structured legal reasoning. 
To further enhance its performance on complex legal tasks, we introduce multi-stage reinforcement learning.
The core objective of this stage is to preserve the rigor of legal reasoning while encouraging the model to develop a more internalized and autonomous ``legal thinking'' pattern. 
This is achieved by carefully designed reward signals that guide the model away from lengthy and loose reasoning processes towards more concise, reliable, and interpretable reasoning structures.

Technically, we employ the DAPO~\cite{yu2025dapo} algorithm as the unified optimization backbone, complemented by a novel token-level baseline designed to minimize variance. The distinction across stages primarily lies in the adaptive context length constraints and the specific design of reward functions. 
As shown in Figure~\ref{figure:RL}, we design five reinforcement learning tasks arranged from easy to hard, which steadily strengthen the three core skill levels in the legal task: memorization, comprehension, and reasoning. 
As the curriculum progresses, the reward signals evolve from fully verifiable objective metrics to rubric-based holistic evaluations. 
In parallel, task difficulty increases along three dimensions: legal complexity, richness of contextual information, and required output length.  Early tasks encourage brief and precise responses, while later tasks mildly reward more detailed and well structured legal arguments. 
Through multi-stage reinforcement learning, the model first consolidates statute-level knowledge, then progressively acquires context-aware norm application, and ultimately develops natural, concise, and trustworthy legal reasoning behavior.

\subsection{Task Defination}

\textbf{Statute Memorization.} 
An accurate command of the statutory text is the prerequisite for trustworthy legal reasoning, as any hallucination at this fundamental level may create errors that propagate throughout the downstream reasoning chain. 
Therefore, the model is firstly required to complete the original text of a provision given only the statute name or the article number. 
Here, we use ROUGE scores as the reward signal to strictly penalize deviations or free-form elaborations. 
Additionally, we impose a maximum generation length of 1,024 tokens. 
This can effectively restrict model generation to authentic statutory language and minimize the risk of possible verbal hallucinations.

\textbf{Statute Application.} 
In addition to memorization, the LLM must possess the ability to map concrete factual descriptions onto abstract legal norms. 
To this end, we aim at teaching the model to identify and exhaustively output all applicable provisions for a given factual scenario.
We utilize the soft match-based F1 score as the reward signal to align the model outputs with the ground truth. 
A generated provision is considered ``matched'' only when the ROUGE~\cite{lin2004rouge} score between it and the target provision exceeds 0.5.
Using this threshold, we calculate Precision and Recall to determine the final F1 score used for optimization. 
Furthermore, we extend the maximum output length to 4,096 tokens. 
This configuration enables the model to exhaustively identify all relevant legal norms in a single pass, ensuring comprehensive alignment with real-world scenarios.

\textbf{Case Analysis.} 
To further enhance the model's ability of analyzing legal facts, procedural contexts, and governing principles, we introduce another reasoning task.  
Adopting a Web-scale data synthesis paradigm, we generate verifiable multiple-choice questions (MCQs) from legal documents. 
Each question features 16 options to eliminate random guessing noise, compelling the model to rely on genuine comprehension. 
We extend the output length to 8,192 tokens and use Accuracy directly as the reward signal. 
This rigorous setup drives a qualitative shift for the model from shallow text matching to deep legal reasoning.

\textbf{Judgment Prediction.}  
Predicting judicial outcomes solely based on case facts requires a complete capability loop—integrating fact comprehension, law application, and judicial discretion, which is quite challenging. 
Rather than mimicking surface text patterns, the model needs to internalize the whole judicial decision-making logic.
We hereby extend the context length to 16k tokens to accommodate detailed case information. 
For the reward mechanism, we employ a ground-truth-based LLM-as-a-judge~\cite{li2024llms,li2025calibraeval}, utilizing distinct evaluation prompts tailored to Civil, Criminal, and Administrative law, respectively. 
To account for authorized judicial discretion in quantitative outcomes, e.g., sentencing terms, fines, and probation periods, we assign the reward to 1 if the predicted value falls within $\pm$10\% of the ground truth, and 0 otherwise.

\textbf{Legal Document Generation.}  
The task involves generating complex legal documents, such as indictments, defense pleadings, and judgments, from given factual descriptions.
It requires the model not only to comprehend facts and apply laws, but also to structure its output in a standardized and legally executable manner.
We adopt a rubric-based evaluation approach, utilizing a dedicated model trained to provide reliable rubrics for various document types. 
The context length is set to 32k tokens.

In summary, these five stages constitute a robust training framework that systematically cultivates the model’s capabilities across the full spectrum of legal cognition, from fundamental memorization to complicated reasoning.

\subsection{Token-Level Baseline Optimization}

\subsubsection{Motivation and Definition}
In the DAPO training process, the estimation of the advantage function is critical for training stability. Conventional approaches typically employ a \textit{Sequence-Level Baseline} ($b_1$), which treats each sample with equal importance. However, we implement a \textit{Token-Level Baseline} ($b_2$) that refines the baseline to token granularity to better account for varying sequence lengths. 

Specifically, for a batch of $n$ samples where each sample $i$ has a length $L_i$ and a sample-level reward $R_i$, every constituent token $j$ is assigned an identical reward $r_{i,j} = R_i$. While the sample-level baseline $b_1$ is the simple arithmetic mean $b_1 = \frac{1}{n} \sum_{i=1}^{n} R_i$, our token-level baseline $b_2$ represents the average reward across all $N$ tokens in the batch ($N = \sum L_i$), effectively functioning as a length-weighted average:
\begin{equation}
b_2 = \frac{\sum_{i=1}^{n} \sum_{j=1}^{L_i} r_{i,j}}{N} = \frac{\sum_{i=1}^{n} L_i R_i}{\sum_{i=1}^{n} L_i}
\end{equation}

This refinement is essential because $b_1$ introduces a distinct length bias. For long negative samples, where $b_1 > b_2$, the absolute value of the negative advantage $|A_{\text{neg}}|$ is artificially inflated, causing the gradient to over-penalize length. Conversely, for long positive samples where $b_2 > b_1$, the increased $|A_{\text{pos}}|$ causes the model to lean excessively toward longer outputs rather than actual quality. By utilizing $b_2$, we ensure that the advantage function remains unbiased while significantly reducing estimation variance.

\subsubsection{Proof of Unbiasedness}
To ensure the theoretical integrity of the training process, we first demonstrate that the introduction of the Token-Level Baseline $b_2$ does not introduce additional bias into the policy gradient. The expected gradient of the objective function with respect to parameters $\theta$ is given by:
\begin{equation}
\mathbf{g} = \mathbb{E}_{\tau \sim \pi_{\theta}} \left[ \sum_{t=1}^{T} \nabla_{\theta} \log \pi_{\theta}(a_t | s_t) (R(\tau) - b) \right]
\end{equation}
Next, we can easily prove that the expected contribution of the baseline term is zero based on the linearity of the expectation:
\begin{equation}
\mathbb{E}_{\tau \sim \pi_{\theta}} \left[ \sum_{t=1}^{T} \nabla_{\theta} \log \pi_{\theta}(a_t | s_t) \cdot b \right] = 0
\end{equation}
For any arbitrary timestep $t$, we apply the Law of Iterated Expectations by conditioning on the trajectory history $h_t = \{s_{1:t}, a_{1:t-1}\}$ prior to the current action:
\begin{equation}
\mathbb{E}_{s_{1:T}, a_{1:T}} \left[ \nabla_{\theta} \log \pi_{\theta}(a_t | s_t) \cdot b \right] = \mathbb{E}_{h_t} \left[ \mathbb{E}_{a_t \sim \pi_{\theta}(\cdot|s_t)} \left[ \nabla_{\theta} \log \pi_{\theta}(a_t | s_t) \cdot b \mid h_t \right] \right]
\end{equation}
Since $b$ is independent of the current action $a_t$ (it is a deterministic statistic of the batch), it can be factored out of the inner expectation:
\begin{equation}
\mathbb{E}_{h_t} \left[ b \cdot \underbrace{\mathbb{E}_{a_t \sim \pi_{\theta}(\cdot|s_t)} \left[ \nabla_{\theta} \log \pi_{\theta}(a_t | s_t) \right]}_{I} \right]
\end{equation}
Using the identity $\nabla \log f = \frac{\nabla f}{f}$, the inner term $I$ becomes:
\begin{equation}
I
= \int \pi_{\theta}(a_t \mid s_t)\,
\frac{\nabla_{\theta}\pi_{\theta}(a_t \mid s_t)}{\pi_{\theta}(a_t \mid s_t)} \, d a_t
= \nabla_{\theta}\int \pi_{\theta}(a_t \mid s_t)\, d a_t
= \nabla_{\theta}(1)
= 0 .
\end{equation}
Thus, the baseline term vanishes in expectation, proving that $b_2$ maintains the unbiasedness of the policy gradient.

\subsubsection{Variance Reduction Analysis}
To demonstrate that the Token-Level Baseline ($b_2$) is superior to the Sample-Level Baseline ($b_1$), we analyze the variance of the gradient estimation. In our context, the variance is proportional to the weighted squared residuals of the rewards: $\text{Var} \propto \sum_{i=1}^n L_i (R_i - b)^2$. We decompose the total variance associated with $b_1$ by centering it around $b_2$:

\begin{equation}
\begin{aligned}
\sum_{i=1}^{n} L_i (R_i - b_1)^2 &= \sum_{i=1}^{n} L_i (R_i - b_2 + b_2 - b_1)^2 \\
&= \sum_{i=1}^{n} L_i (R_i - b_2)^2 + \sum_{i=1}^{n} L_i (b_2 - b_1)^2 + 2 \sum_{i=1}^{n} L_i (R_i - b_2)(b_2 - b_1)
\end{aligned}
\end{equation}

Consider the cross-product term. Since $(b_2 - b_1)$ is a constant scalar relative to the summation over $i$, it can be factored out:
\begin{equation}
2(b_2 - b_1) \sum_{i=1}^{n} L_i (R_i - b_2) = 2(b_2 - b_1) \left( \sum_{i=1}^{n} L_i R_i - b_2 \sum_{i=1}^{n} L_i \right)
\end{equation}
Recalling $b_2 = \frac{\sum L_i R_i}{N}$, the term inside the parentheses is $b_2 N - b_2 N = 0$. Consequently, the decomposition simplifies to:
\begin{equation}
\underbrace{\sum_{i=1}^{n} L_i (R_i - b_1)^2}_{\text{Var}_1 \text{ (Total Variance)}} = \underbrace{\sum_{i=1}^{n} L_i (R_i - b_2)^2}_{\text{Var}_2 \text{ (Minimal Variance)}} + \underbrace{N(b_2 - b_1)^2}_{\text{Variance Gap } \ge 0}
\end{equation}
The derivation explicitly shows that $\text{Var}_1 \ge \text{Var}_2$. The Variance Gap represents the additional noise introduced by the correlation between sequence length and reward. By minimizing this variance, $b_2$ ensures a more stable optimization landscape for the LegalOne model.

\section{Evaluation}

\subsection{EXPERIMENTAL SETUP}
To facilitate result reproducibility, we introduce LegalKit, a practical and scalable toolkit designed for evaluating Large Language Models (LLMs) in the legal domain. We utilize LegalKit to conduct a comprehensive and meticulous assessment. For the evaluation settings, we configure the temperature to 0.6 and the maximum context length to 16k tokens.
We evaluate LLMs using two of the most prominent and comprehensive benchmarks designed for the legal domain:

\begin{itemize}

\item \textbf{JEC-QA~\cite{zhong2020jec}.} Sourced from the National Unified Legal Professional Qualification Examination, JEC-QA is the largest question-answering dataset in the legal field.  he dataset categorizes queries into Knowledge-Driven questions (KD-questions) and Case-Analysis questions (CA-questions).

\item \textbf{LexEval~\cite{li2024lexeval}.} LexEval proposes a novel Legal Cognitive Ability Taxonomy to systematically organize various legal tasks. This taxonomy encompasses six core abilities: Memorization, Understanding, Logical Inference, Discrimination, Generation, and Ethics. Notably, while the original LexEval benchmark typically relies on ROUGE metrics for Generation tasks, we adopt Qwen3-235B-A22B~\cite{yang2025qwen3} as an LLM-as-a-Judge to ensure a more robust and reliable evaluation. The specific prompts used in our experiments are accessible on GitHub.


\end{itemize} 

\begin{figure*}[t]
\centering
\includegraphics[width=0.8\linewidth]{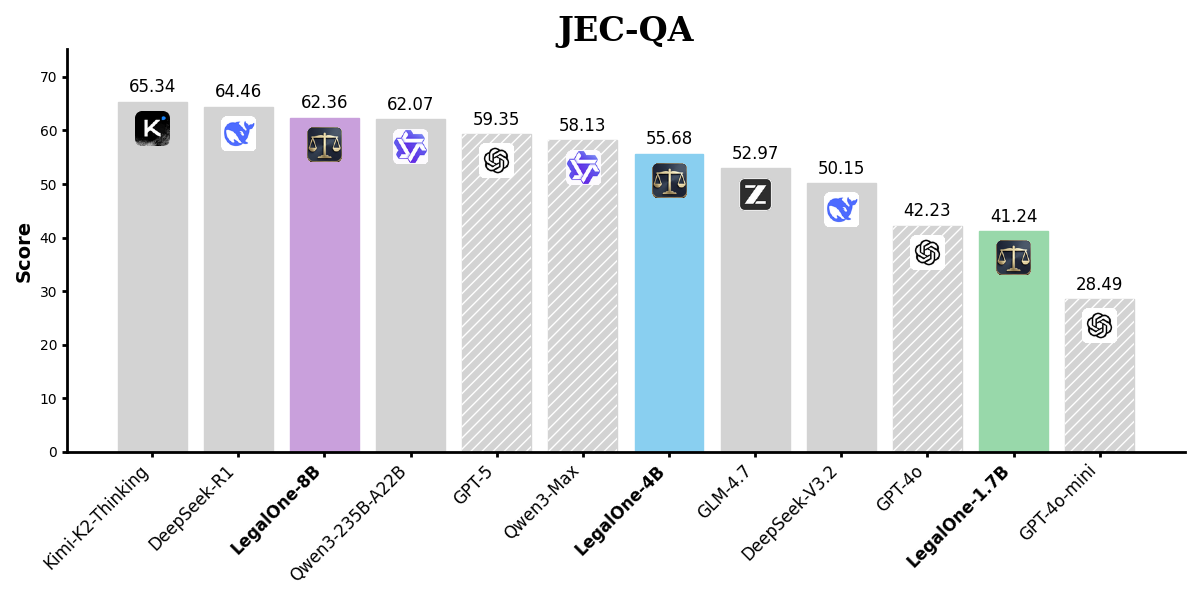}
\caption{Evaluation results of various LLMs on the JEC-QA benchmark.}
\label{figure:jecqa}
\end{figure*}

\begin{figure*}[t]
\centering
\includegraphics[width=0.8\linewidth]{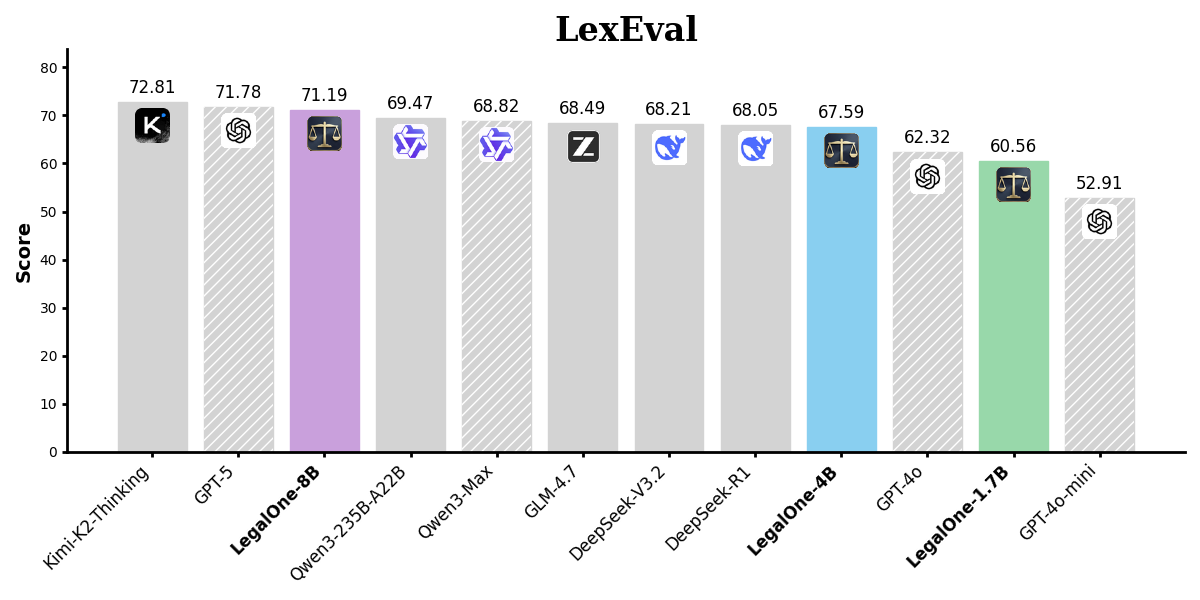}
\caption{Evaluation results of various LLMs on the LexEval benchmark.}
\label{figure:lexeval}
\end{figure*}

\subsubsection{Comparison Models}
We developed three models of varying scales, LegalOne-8B, LegalOne-4B, and LegalOne-1.7B, all initialized from the Qwen3-base. These models were benchmarked against eight representative, state-of-the-art LLMs: Qwen3-235B-A22B-Thinking~\cite{yang2025qwen3}, Qwen3-Max~\cite{yang2025qwen3}, GLM-4.7~\cite{zeng2025glm}, DeepSeek-R1~\cite{guo2025deepseek}, DeepSeek-V3.2~\cite{liu2025deepseek}, Kimi-K2-thinking~\cite{team2025kimi}, GPT-5~\cite{singh2025openai}, GPT-4o~\cite{singh2025openai}, and GPT-4o-mini~\cite{singh2025openai}. These models represent the current frontiers of general domain LLM with robust and widely validated performance. Notably, we excluded existing legal domain LLMs from our baselines, as their overall capabilities currently lag significantly behind those of leading general-domain models~\cite{li2025lexrag,li2025casegen}.
\subsection{MAIN RESULTS}
Figure \ref{figure:jecqa} and Figure \ref{figure:lexeval} illustrate the overall performance rankings across the JEC-QA and LexEval benchmarks, respectively. Notably, LegalOne-8B, with a significantly smaller parameter footprint, achieves performance comparable to much larger general-purpose models. Detailed numerical results for each evaluation dimension are presented in Table \ref{tab:model_performance_detailed}, while the full task-specific performance breakdown for LexEval is provided in the Appendix~\ref{exp}. Our analysis reveals several key insights:

\begin{table}[t]
    \centering
    \caption{Model Performance Comparison on LexEval and JEC-QA Benchmarks. The \textbf{bold} values indicate the best performance in each column, while the \underline{underlined} values represent the second-best performance.}
    \label{tab:model_performance_detailed}
    \begin{adjustbox}{width=\textwidth}
    \begin{tabular}{lcccccccccc}
        \toprule
        \multirow{2}{*}{\textbf{Model}} & \multicolumn{7}{c}{\textbf{LexEval}} & \multicolumn{3}{c}{\textbf{JEC-QA}} \\
        \cmidrule(lr){2-8} \cmidrule(lr){9-11}
        & \textbf{Memo.} & \textbf{Under.} & \textbf{Logic} & \textbf{Disc.} & \textbf{Gen.} & \textbf{Ethic} & \textbf{Avg.} & \textbf{KD} & \textbf{CA} & \textbf{Avg.} \\
        \midrule
        GPT-4o-mini      & 24.19 & 83.59 & 50.23 & 30.90 & 59.03 & 42.40 & 52.91 & 29.46 & 27.52 & 28.49 \\
        GPT-4o           & 39.74 & 84.12 & 65.77 & 33.97 & 63.93 & 58.47 & 62.32 & 46.27 & 38.18 & 42.23 \\
        GPT-5            & 62.31 & 85.69 & 73.59 & 41.07 & \underline{72.88} & \textbf{73.43} & \underline{71.78} & 60.88 & 57.83 & 59.35 \\
        Qwen3-235B-A22B  & 58.32 & 85.61 & 73.24 & 37.47 & \textbf{73.50} & 62.17 & 69.47 & 65.16 & 58.98 & 62.07 \\
        Qwen3-Max        & 58.46 & 88.00 & \underline{75.82} & 33.00 & 66.88 & 59.70 & 68.82 & 62.03 & 54.22 & 58.13 \\
        DeepSeek-v3.2    & 55.40 & 86.67 & 73.19 & 29.60 & 70.20 & 63.37 & 68.21 & 54.58 & 45.72 & 50.15 \\
        DeepSeek-R1      & 49.64 & 86.91 & 74.07 & 33.80 & 68.88 & 64.77 & 68.05 & \underline{67.88} & 61.04 & \underline{64.46} \\
        GLM-4.7          & 56.50 & \underline{88.31} & 72.86 & 33.70 & 69.18 & 61.03 & 68.49 & 56.90 & 49.05 & 52.97 \\
        Kimi-K2-thinking & \textbf{63.78} & \textbf{88.75} & \textbf{76.41} & \textbf{42.90} & 72.38 & \underline{68.60} & \textbf{72.81} & \textbf{69.49} & \underline{61.19} & \textbf{65.34} \\
        \midrule
        Qwen3-1.7B       & 29.66 & 64.29 & 54.93 & 24.34 & 51.50 & 48.23 & 49.54 & 31.02 & 25.25 & 28.13 \\
        Qwen3-4B         & 44.26 & 78.49 & 54.74 & 23.54 & 59.15 & 55.63 & 56.71 & 41.99 & 37.27 & 39.63 \\
        Qwen3-8B         & 49.62 & 79.31 & 60.16 & 29.77 & 60.65 & 57.60 & 60.06 & 48.44 & 40.78 & 44.61 \\
        LegalOne-1.7B    & 50.96 & 82.55 & 65.63 & 30.63 & 55.63 & 49.93 & 60.56 & 41.94 & 40.54 & 41.24 \\
        LegalOne-4B      & 58.59 & 86.63 & 72.62 & 36.87 & 63.48 & 60.77 & 67.59 & 58.41 & 52.95 & 55.68 \\
        LegalOne-8B      & \underline{62.68} & 87.77 & 74.66 & \underline{41.63} & 68.95 & 67.83 & 71.19 & 63.19 & \textbf{61.52} & 62.36 \\
        \bottomrule
    \end{tabular}
    \end{adjustbox}
\label{table:overall}
\end{table}

\begin{itemize}

\item While experimental results generally align with scaling laws, evidenced by Kimi-K2-thinking (1T parameters) achieving SOTA performance, the result reveals that domain-specific high-quality data and targeted training methodologies act as catalysts for increasing  LLM's ``capability density''. Specifically, the LegalOne-1.7B significantly outperforms the general-purpose Qwen3-1.7B and delivers performance comparable to the much larger Qwen3-8B. It demonstrates that maximizing capability density through curated data is a viable pathway to bridge the performance gap, allowing compact models to handle intricate professional tasks previously reserved for trillion-parameter giants.

\item LegalOne demonstrates exceptional parameter efficiency. Despite its compact size, LegalOne-8B delivers highly competitive performance, surpassing significantly larger models such as DeepSeek-v3.2 and Qwen3-Max in multiple dimensions. Notably, it secures the top score in JEC-QA CA (61.52) and ranks second in LexEval’s Discrimination and Ethics metrics. The success of the LegalOne series offers a robust blueprint for developing small-yet-reliable models in specialized fields, paving the way for the deployment of high-performance, privacy-compliant, and cost-effective AI in resource-constrained professional environments.

\end{itemize} 

\subsection{Ablation Study}

\begin{table*}[t]
\centering
\small 
\setlength{\tabcolsep}{12pt} 
\caption{Ablation study of LegalOne. The results demonstrate the effectiveness of our method.}
\label{tab:ablation_study}
\begin{tabular}{lcccccc}
\toprule
\multirow{2}{*}{\textbf{Method}} & \multicolumn{3}{c}{\textbf{JEC-QA}} & \multicolumn{3}{c}{\textbf{LexEval}} \\
\cmidrule(lr){2-4} \cmidrule(lr){5-7}
 & \textbf{1.7B} & \textbf{4B} & \textbf{8B} & \textbf{1.7B} & \textbf{4B} & \textbf{8B} \\
\midrule
\multicolumn{7}{l}{\textit{Baseline (Official)}} \\
Qwen3-Thinking & 28.13 & 39.63 & 44.61 & 49.54 & 56.71 & 60.06 \\
\midrule
\multicolumn{7}{l}{\textit{Ours (from Qwen3-Base)}} \\
\hspace{0.3cm} w SFT & 26.26 & 46.02 & 52.56 & 44.77 & 59.69 & 63.68 \\
\hspace{0.3cm} w MT + SFT & 39.78 & 53.83 & 56.37 & 57.07 & 66.11 & 68.32 \\
\hspace{0.3cm} w MT + SFT + RL & \textbf{41.24} & \textbf{55.68} & \textbf{62.36} & \textbf{60.56} & \textbf{67.59} & \textbf{71.19} \\
\bottomrule
\end{tabular}
\label{tab:ablation_study}
\vspace{-5mm}
\end{table*}

To validate the efficacy of our training pipeline, we conducted a progressive ablation study across varying model scales (1.7B, 4B, and 8B). All experimental variants were initialized from Qwen3-Base, with the official post-trained Qwen3-Thinking serving as the comparative baseline. The results, summarized in Table \ref{tab:ablation_study}, yield the following key insights:

\begin{itemize}

\item On the 4B and 8B scales, applying SFT using the LEAD pipeline alone was sufficient to surpass the official baseline, demonstrating the effectiveness of transforming judicial processes into structured reasoning trajectories. However, the 1.7B model exhibited performance degradation (falling below the baseline) during the SFT phase. We guess this may be due to the lack of domain priors in smaller models, resulting in poor robustness when facing complex legal instructions. This observation underscores the necessity of Mid-training, confirming that domain competence is a prerequisite for effective instruction alignment.

\item By establishing a high-density legal semantic space, Mid-training drove a qualitative performance leap, particularly enabling the 1.7B model to outperform the baseline. We observed that SFT converges more stably after Mid-training. This suggests that the injected domain knowledge acts as a stabilizer, optimizing the loss landscape for subsequent alignment tasks.

\item Curriculum RL further propels the LLM's reasoning capabilities. Notably, we observed a significant trend where larger LLMs (e.g., 8B) achieved substantially higher performance gains during the RL phase. We attribute this to the superior policy optimization potential inherent in larger LLMs, which enables more effective exploration of the solution space. This finding suggests the existence of a scaling Law specific to legal reasoning within the RL stage, providing empirical evidence for future model scaling.

\end{itemize}

\subsection{General Capability}

In this section, we evaluate the LLM's general capabilities in non-legal domains using the GSM8K~\cite{cobbe2021gsm8k}, Math500~\cite{lightman2023lets}, and C-Eval~\cite{huang2023c} benchmarks. It is worth noting that, due to computational constraints, we did not incorporate large-scale general domain replay data during the SFT and RL stages to explicitly preserve or activate these capabilities. Consequently, our analysis primarily focuses on the evolutionary trends of the Qwen3-Base model across different training stages. As shown in Table~\ref{tab:general}, the experimental results yield three key conclusions:

(1). First, performance in general domains adheres to the Scaling Law. As model parameters increase from 1.7B to 8B, all general metrics exhibit a consistent upward trend. This validates that larger model capacity provides a more robust foundation for cross-domain knowledge transfer.

(2). Second, MT+SFT typically performs better than direct SFT. While direct SFT risks disrupting parameter distributions and causing catastrophic forgetting, MT acts as a bridge between the base model and downstream tasks. It injects domain knowledge while preserving fundamental language skills. Crucially, the stability of these general capabilities validates the effectiveness of our PAS sampling algorithm in mitigating forgetting.

(3). Notably, we observe that mid-training may lead to a slight degradation in general capabilities for smaller models (e.g., 1.7B). We attribute this to the ``capacity bottleneck'', where limited parameter space struggles to accommodate both new domain knowledge and existing general knowledge simultaneously. Consequently, we recommend that domain-specific model training be conducted on models with sufficient parameter scale to mitigate such trade-offs.

(4). Finally, the most surprising finding is the positive transfer of RL to general capabilities. Despite using exclusively legal data for RL training, the model's general reasoning improved unexpectedly. We attribute this to the shared nature of logical reasoning. Legal tasks require rigorous logic, such as syllogisms and multi-step analysis. By mastering these strict thought patterns through RL, the model strengthens its core reasoning capabilities and generalizes to other domains. This inspiring finding motivates future research into how domain-specific RL can enhance general intelligence by refining fundamental reasoning skills.

In summary, the LegalOne series preserves strong general capabilities, validating our method's ability to inject professional knowledge without sacrificing the model's foundational intelligence.

\begin{table}[t!]
    \centering
    \caption{Performance comparison of Qwen3-Base models across different training stages on GSM8K, Math500, and C-Eval benchmarks. The best results in each category are highlighted in bold.}
    \label{tab:general}
    \setlength{\tabcolsep}{4pt} 
    \begin{tabular}{@{}lccccccccc@{}}
        \toprule
        \multirow{2}{*}{\textbf{Method}} & \multicolumn{3}{c}{\textbf{GSM8K}} & \multicolumn{3}{c}{\textbf{Math500}} & \multicolumn{3}{c}{\textbf{C-Eval}} \\
        \cmidrule(lr){2-4} \cmidrule(lr){5-7} \cmidrule(lr){8-10}
         & \textbf{1.7B} & \textbf{4B} & \textbf{8B} & \textbf{1.7B} & \textbf{4B} & \textbf{8B} & \textbf{1.7B} & \textbf{4B} & \textbf{8B} \\
        \midrule
        w SFT        & \textbf{62.62} & 74.67 & 84.0 & 24.6 & 33.4 & 60.6 & 45.00 & 41.65 & 69.28 \\
        w MT+SFT    & 48.98 & 77.71 & 83.7 & 26.8 & 61.0 & 65.6 & 38.80 & 59.43 & 65.04 \\
        w MT+SFT+RL & 49.58 & \textbf{81.43} & \textbf{85.9} & \textbf{33.6} & \textbf{63.8} & \textbf{73.2} & \textbf{46.72} & \textbf{62.07} & \textbf{70.65} \\
        \bottomrule
    \end{tabular}
\end{table}

\subsection{Case Study}

Table \ref{tab:legalone_case} presents a representative example of LegalOne conducting legal reasoning. In this scenario, the model is tasked with analyzing a complex criminal case and predicting the court's judgment. 

As illustrated, LegalOne exhibits a highly structured reasoning process. It begins by performing a comprehensive fact analysis to identify key evidentiary details. Subsequently, it systematically applies the \textit{Three-Tier Theory of Crime}, analyzing Constitutive Requirements, Illegality, and Culpability, to establish criminal liability. Finally, the model proceeds to sentencing analysis, explicitly citing relevant legal statutes to support its arguments at each stage.

This structured chain of thought closely mirrors the cognitive patterns found in professional legal practice, a capability that was notably absent in previous LLMs. We argue that this sophisticated behavior is a direct result of our proposed LEAD framework combined with Reinforcement Learning. This effective integration allows LegalOne to bridge the gap between AI generation and professional legal standards, highlighting its potential to provide substantial assistance to legal practitioners.

\begin{table}[t!]
    \centering
    \caption{An example of LegalOne conducting legal reasoning.}
    \label{tab:legalone_case}
    \begin{tabular}{@{} p{0.18\textwidth} p{0.78\textwidth} @{}}
        \toprule
        \textbf{Role} & \textbf{Content} \\
        \midrule
        
        \textbf{Query} & 
        Please analyze the following case facts from a legal perspective and predict the court's judgment. \newline
        Case Facts: \newline
        Since June 2017, Defendant Qiu Shouqiang has engaged in online gambling and became the General Agent for the gambling website ``CC Color Ball.'' After obtaining the General Agency rights for the site, Qiu Shouqiang conspired with his wife, Defendant Cheng Jinling, to obtain rebates and dividends by recruiting subordinate agents and members to gamble on the website. Within this operation, Qiu Shouqiang was primarily responsible for developing lower-level agents and members, while Cheng Jinling, acting under the name ``XXX Finance,'' was primarily responsible for managing the crediting and debiting of points (deposits and withdrawals)... \\
        
        \midrule 
        
        \textbf{LegalOne-8B} & 
        Based on the provided case facts, I will conduct a legal analysis and predict the court's potential judgment. I will apply the Three-Tier Theory of Crime (Constitutive Requirements, Illegality, and Culpability) to analyze the case and incorporate sentencing guidelines to forecast the verdict. \newline
        First, I will outline the facts of the case: ... \newline
        ~ \newline
        Now, I will proceed with the analysis based on the Three-Tier Theory of Crime: \newline
    
      1. Objective Elements According to Article 303, Paragraph 2 of the Criminal Law of the People's Republic of China, the crime of Opening a Casino refers to the act of opening a casino or acting as an agent for a gambling website and accepting bets for the purpose of profit. Regarding the facts of this case: ... \newline
        ~ \newline
        2. Subjective Elements The crime of Opening a Casino requires the perpetrator to have criminal intent (mens rea) and the purpose of profit. Regarding the facts of this case: ... \newline
        ~ \newline
        3. Culpability Culpability refers to whether the perpetrator should bear criminal responsibility for their unlawful acts. Regarding the facts of this case: ... \newline
        ~ \newline
        Sentencing Analysis \newline
        According to Article 303, Paragraph 2 of the Criminal Law of the PRC, whoever opens a casino shall be sentenced to fixed-term imprisonment of not more than five years, criminal detention, or public surveillance, and shall also be fined; if the circumstances are serious, they shall be sentenced to fixed-term imprisonment of not less than five years but not more than ten years, and shall also be fined... \\
        \bottomrule
    \end{tabular}
\end{table}

\section{Related Works}
\subsection{Legal-Domain Large Language Models}
The increasingly sophisticated capabilities of LLMs drive a paradigm shift in legal domain. 
While general-purpose LLMs like GPT~\cite{openai2025gpt5} and Gemini~\cite{team2023gemini} exhibit remarkable capabilities, they often lack the specialized knowledge required for authoritative legal tasks and frequently hallucinate. 
To address these limitations, numerous domain-specific LLMs has emerged. 
In the Chinese legal context, ChatLaw~\cite{cui2023chatlaw} incorporates knowledge graphs to provide grounded legal information, effectively reducing hallucinations and improving the accuracy of legal consultations.
Similarly, LawGPT~\cite{zhou2024lawgpt} and Fuzi-Mingcha~\cite{sdu2023fuzimingcha} employ SFT on large-scale legal dialogue and judgment datasets to improve consultation capabilities.
In the English legal context, SaulLM~\cite{colombo2024saullm} serves as a major baseline, demonstrating superior performance in text comprehension through the use of legal-specific data. 
However, despite these advancements, most existing legal LLMs often struggle to exhibit the reasoning capabilities required for complex statutory language and legal logic~\cite{LexiLaw,li2025blade}.

\subsection{Mid-Training and Domain Adaptation}
To bridge the gap between general pre-training and domain-specific application, recent research emphasizes \emph{mid-training} as a critical phase for injecting high-density expert knowledge. 
Unlike standard SFT, mid-training involves continued pre-training on massive, unlabeled domain-specific corpora to align the LLMs’ internal knowledge with legal syntax and vocabulary. 
For instance, SaulLM~\cite{colombo2024saullm} was trained on a legal corpus of over 30 billion, significantly outperforming general LLMs on legal benchmarks. 
LawGPT~\cite{zhou2024lawgpt} similarly constructed a pre-training corpus of 500k legal documents to reinforce foundational legal understanding.
However, a major challenge in this phase is \textbf{catastrophic forgetting}, where the injection of intensive legal data degrades the LLMs' general capabilities.

\subsection{Reinforcement Learning for Legal Reasoning}
RL has become pivotal in enhancing the reasoning behaviors of LLMs.
Following the success of DeepSeek-R1~\cite{deepseek2025r1}, which demonstrated that RL could induce long chain-of-thought behaviors, legal researchers have adapted these techniques to enforce judicial logic. 
A significant trend is the shift from Proximal Policy Optimization (PPO)~\cite{schulman2017ppo} to Group Relative Policy Optimization (GRPO)~\cite{deepseek2025r1,shao2024deepseekmath}, which eliminates the need for a critic model and improves training stability.
Legal$\Delta$~\cite{dai2025legaldelta} introduces an information-gain-enhanced reward mechanism within the RLVR (Reinforcement Learning with Verifiable Rewards) framework, encouraging LLMs to explore diverse reasoning trajectories that maximize confidence differentials between direct and reasoning-augmented outputs. 
Furthermore, Unilaw-R1~\cite{cai2025unilaw} transitions from rule-based rewards for verifiable tasks to holistic evaluations for complex reasoning.
Existing RL frameworks for legal domain primarily focus on surface-level verification, frequently overlooking the underlying logical consistency of the entire reasoning chain.
Consequently, bridging the gap between mere answer correctness and the internalization of a rigorous ``Legal Mentality'' remains a critical frontier for future research.

\section{Conclusion}
In this work, we presented LegalOne, a family of foundation models explicitly engineered to bridge the gap between general-purpose linguistic fluency and the rigorous demands of professional legal reasoning. 
To overcome the challenges where general LLMs are hindered by a lack of precise domain knowledge and the complexity of performing rigorous multi-step judicial reasoning, we introduced a comprehensive three-phase training pipeline.
Our proposed Plasticity-Adjusted Sampling (PAS) effectively mitigates catastrophic forgetting during massive knowledge injection, while Legal Agentic CoT Distillation (LEAD) successfully translates implicit judicial logic into explicit, learnable reasoning trajectories. 
Furthermore, our multi-stage Reinforcement Learning curriculum enables the model to evolve from simple pattern matching to autonomous and reliable legal reasoning.
Empirical results demonstrate that LegalOne significantly outperforms existing baselines across a diverse array of legal tasks, setting a new state-of-the-art for Chinese legal LLMs. 
Beyond algorithmic advancements, the release of the LegalOne model checkpoints and the LegalKit evaluation framework serves as a critical infrastructure contribution, promoting reproducibility and lowering barriers for future research. 
We hope LegalOne paves the way for the deployment of trustworthy and interpretable foundation models in high-stakes judicial applications.

\section*{Acknowledgments}
We gratefully acknowledge the Huawei ``Hundred Schools Program'' for providing Ascend computing resources. This work was supported by the Research Project of Quan Cheng Laboratory, China (Grant No.~QCL20250105).

\clearpage
\bibliographystyle{plain}
\bibliography{iclr2026_conference.bib}
\clearpage

\appendix

\section{Detailed Experimental Results}
\label{exp}

\begin{table}[htbp]
\centering
\caption{Zero-shot performance(\%) of various models at Memorization, Understanding, and Logic Inference level on LexEval.}
\label{tab:lex_1}
\resizebox{\textwidth}{!}{%
\begin{tabular}{lcccccccccccccc}
\toprule
\multirow{2.5}{*}{Model} & \multicolumn{3}{c}{Memorization} & \multicolumn{5}{c}{Understanding} & \multicolumn{6}{c}{Logic Inference} \\
\cmidrule(lr){2-4} \cmidrule(lr){5-9} \cmidrule(lr){10-15} 
& 1-1 & 1-2 & 1-3 & 2-1 & 2-2 & 2-3 & 2-4 & 2-5 & 3-1 & 3-2 & 3-3 & 3-4 & 3-5 & 3-6 \\
\midrule
GPT-4o-mini & 27.40 & 34.50 & 10.67 & 86.40 & 52.33 & 93.00 & 89.00 & 97.20 & 79.70 & 68.70 & 25.40 & 26.20 & 48.75 & 52.60 \\
GPT-4o & 37.80 & 68.10 & 13.33 & 92.00 & 57.00 & 92.00 & 82.00 & 97.60 & 81.10 & 93.40 & 58.20 & 35.40 & 62.50 & 64.00 \\
GPT-5 & 59.60 & 96.00 & \textbf{31.33} & 91.60 & 49.67 & 92.00 & 95.80 & 99.40 & 71.60 & 89.40 & 70.30 & 54.20 & \textbf{92.25} & 63.80 \\
Qwen3-235B & 61.40 & 95.90 & 17.67 & 87.80 & 52.67 & 95.00 & 95.20 & 97.40 & 73.90 & 95.40 & 66.00 & 56.20 & 90.75 & 57.20 \\
Qwen3-Max & 57.40 & 99.30 & 18.67 & \textbf{95.80} & 60.00 & 95.00 & 91.00 & 98.20 & 80.60 & 96.80 & \textbf{72.80} & 53.40 & 86.50 & 64.80 \\
DeepSeek-v3.2 & 57.60 & 91.60 & 17.00 & 93.40 & 54.33 & 96.00 & 92.00 & 97.60 & 81.30 & 97.20 & 63.50 & 53.80 & 85.75 & 57.60 \\
DeepSeek-R1 & 57.60 & 72.00 & 19.33 & 92.80 & 59.33 & 96.00 & 90.00 & 96.40 & 83.10 & 96.80 & 59.70 & 55.80 & 92.00 & 57.00 \\
GLM-4.7 & 51.80 & 96.70 & 21.00 & 94.00 & \textbf{60.33} & 95.00 & 95.80 & 96.40 & 80.60 & 97.40 & 67.30 & 47.60 & 80.25 & 64.00 \\
Kimi-K2-thinking & \textbf{66.40} & 98.60 & 26.33 & 91.80 & 59.33 & 97.00 & 95.80 & \textbf{99.80} & \textbf{81.30} & 97.60 & 61.20 & \textbf{59.60} & 89.75 & \textbf{69.00} \\
\midrule 
Qwen3-1.7B & 26.20 & 49.10 & 13.67 & 73.80 & 34.67 & 88.00 & 56.20 & 68.80 & 77.80 & 83.30 & 26.30 & 25.40 & 72.00 & 44.80 \\
Qwen3-4B & 33.80 & 81.30 & 17.67 & 82.60 & 40.67 & 94.00 & 81.80 & 93.40 & 80.30 & 87.30 & 15.40 & 25.40 & 78.25 & 41.80 \\
Qwen3-8B & 40.40 & 87.80 & 20.67 & 86.80 & 46.33 & 94.00 & 73.20 & 96.20 & 80.70 & 91.00 & 34.20 & 30.20 & 78.25 & 46.60 \\
LegalOne-1.7B & 40.20 & 96.00 & 16.67 & 84.20 & 49.33 & 92.00 & 91.40 & 95.80 & 79.20 & 94.60 & 51.40 & 41.80 & 74.00 & 52.80 \\
LegalOne-4B & 55.40 & 99.70 & 20.67 & 87.20 & 54.33 &\textbf{97.00} & 96.40 & 98.20 & 80.70 & 97.10 & 60.80 & 52.80 & 86.50 & 57.80 \\
LegalOne-8B & 60.00 & \textbf{99.70} & 28.33 & 91.60 & 55.67 & 95.00 & \textbf{97.60} & 99.00 & 80.10 & \textbf{98.00} & 57.60 & 59.40 & 89.25 & 63.60 \\
\bottomrule
\end{tabular}%
}
\end{table}

\begin{table}[htbp]
\centering
\caption{Zero-shot performance(\%) of various models at Discrimination, Generation, and Ethic level on LexEval..}
\label{tab:lex_2}
\resizebox{\textwidth}{!}{%
\begin{tabular}{lcccccccccc}
\toprule
\multirow{2.5}{*}{Model} & \multicolumn{2}{c}{Discrimination} & \multicolumn{4}{c}{Generation} & \multicolumn{3}{c}{Ethic} & \multirow{2.5}{*}{Average} \\
\cmidrule(lr){2-3} \cmidrule(lr){4-7} \cmidrule(lr){8-10}
& 4-1 & 4-2 & 5-1 & 5-2 & 5-3 & 5-4 & 6-1 & 6-2 & 6-3 & \\
\midrule
GPT-4o-mini & 29.80 & 32.00 & 68.10 & 46.40 & 71.40 & 50.20 & 27.40 & 38.20 & 61.60 & 52.91 \\
GPT-4o & 31.60 & 36.33 & 71.50 & 51.40 & 71.80 & 61.00 & 49.80 & 55.00 & 70.60 & 62.32 \\
GPT-5 & \textbf{38.80} & 43.33 & \textbf{76.40} & 59.30 & \textbf{78.10} & 77.70 & \textbf{69.20} & 66.50 & \textbf{84.60} & 71.78 \\
Qwen3-235B & 32.60 & 42.33 & 74.80 & \textbf{60.70} & 77.50 & \textbf{81.00} & 51.40 & 62.30 & 72.80 & 69.47 \\
Qwen3-Max & 22.00 & 44.00 & 73.70 & 58.60 & 75.10 & 60.10 & 42.20 & 60.70 & 76.20 & 68.82 \\
DeepSeek-v3.2 & 21.20 & 38.00 & 73.10 & 57.60 & 75.20 & 74.90 & 48.50 & 62.80 & 78.80 & 68.21 \\
DeepSeek-R1 & 25.60 & 42.00 & 74.10 & 55.30 & 74.40 & 71.70 & 51.40 & 60.30 & 82.60 & 68.05 \\
GLM-4.7 & 26.40 & 41.00 & 71.80 & 55.60 & 75.80 & 73.50 & 41.20 & 62.50 & 79.40 & 68.49 \\
Kimi-K2-thinking & 36.80 & \textbf{49.00} & 74.60 & 59.20 & 76.70 & 79.00 & 59.90 & \textbf{66.70} & 79.20 & \textbf{72.81} \\
\midrule 
Qwen3-1.7B & 25.00 & 23.67 & 58.90 & 40.90 & 61.30 & 44.90 & 35.70 & 43.20 & 65.80 & 49.54 \\
Qwen3-4B & 14.40 & 32.67 & 67.30 & 48.50 & 68.40 & 52.40 & 48.90 & 49.00 & 69.00 & 56.71 \\
Qwen3-8B & 21.20 & 38.33 & 68.50 & 48.40 & 70.00 & 55.70 & 48.10 & 54.10 & 70.60 & 60.06 \\
LegalOne-1.7B & 29.60 & 31.67 & 63.00 & 52.90 & 44.40 & 62.20 & 40.00 & 50.20 & 59.60 & 60.56 \\
LegalOne-4B & 30.40 & 43.33 & 68.40 & 53.40 & 62.10 & 70.00 & 52.70 & 61.80 & 67.80 & 67.59 \\
LegalOne-8B & 37.60 & 45.67 & 74.00 & 57.70 & 68.10 & 76.00 & 61.80 & 65.70 & 76.00 & 71.19 \\
\bottomrule
\end{tabular}%
}
\end{table}

\end{document}